\documentclass[11pt]{article}
\usepackage{coling2020}
\usepackage{times}
\usepackage{url}
\usepackage{latexsym}
\usepackage{microtype}
\usepackage{booktabs}       
\usepackage{amsfonts}
\usepackage{amsmath}
\usepackage{algorithm}
\usepackage{algorithmic}
\usepackage{multirow}
\usepackage{graphics}
\usepackage{graphicx}
\usepackage{float}
\usepackage{wrapfig}
\usepackage{lipsum}
\usepackage{color}

\colingfinalcopy 

\title{Connecting the Dots Between Fact Verification and Fake News Detection}

\author{
Qifei Li\thanks{\ \ Equal Contribution.} ~~~ Wangchunshu Zhou$^*$ \\
Beihang University\\
{\tt \{liqifei, zhouwangchunshu\}@buaa.edu.cn}}

\begin{document}
\maketitle
\begin{abstract}
Fact verification models have enjoyed a fast advancement in the last two years with the development of pre-trained language models like BERT and the release of large scale datasets such as FEVER. However, the challenging problem of fake news detection has not benefited from the improvement of fact verification models, which is closely related to fake news detection. In this paper, we propose a simple yet effective approach to connect the dots between fact verification and fake news detection. Our approach first employs a text summarization model pre-trained on news corpora to summarize the long news article into a short claim. Then we use a fact verification model pre-trained on the FEVER dataset to detect whether the input news article is real or fake. Our approach makes use of the recent success of fact verification models and enables zero-shot fake news detection, alleviating the need of large scale training data to train fake news detection models. Experimental results on FakenewsNet, a benchmark dataset for fake news detection, demonstrate the effectiveness of our proposed approach.
\end{abstract}

\section{Introduction}

Recently, fake news has been appearing in large numbers, which are easily accessed and disseminated in the online world with the booming development of online social networks. Users can be affected due to the deceptive words intentionally and verifiably false, which makes fake news detection urgent and important for maintaining social order~\cite{shu2017fake}. 

Most existing methods for detecting fake news rely heavily on supervised learning on a large scale dataset with news articles labeled as fake or real by human experts. However, such a labeled dataset is difficult and time-consuming to obtain while few large scale fake news detection datasets are publicly available~\cite{oshikawa2018survey}. Meanwhile, with the fast development of pre-trained language models~\cite{peters2018deep,radford2018improving,devlin2018bert} and the release of large scale datasets~\cite{thorne2018fever}, research on the fact verification task have enjoyed fast advancement~\cite{nie2019combining,hanselowski2018ukp,zhou2019gear}. However, the advancement in the field of fact verification fails to transfer to the task of fake news detection due to the different nature of input sequences (i.e., fact verification aims to check the reliability of a claim of one or a few sentences while fake news detection aims to check the trustworthy of a long article) and the lack of large scale training data to train the fact verification models for fake news detection.

\begin{figure*}
    \centering
    \includegraphics[width=\textwidth]{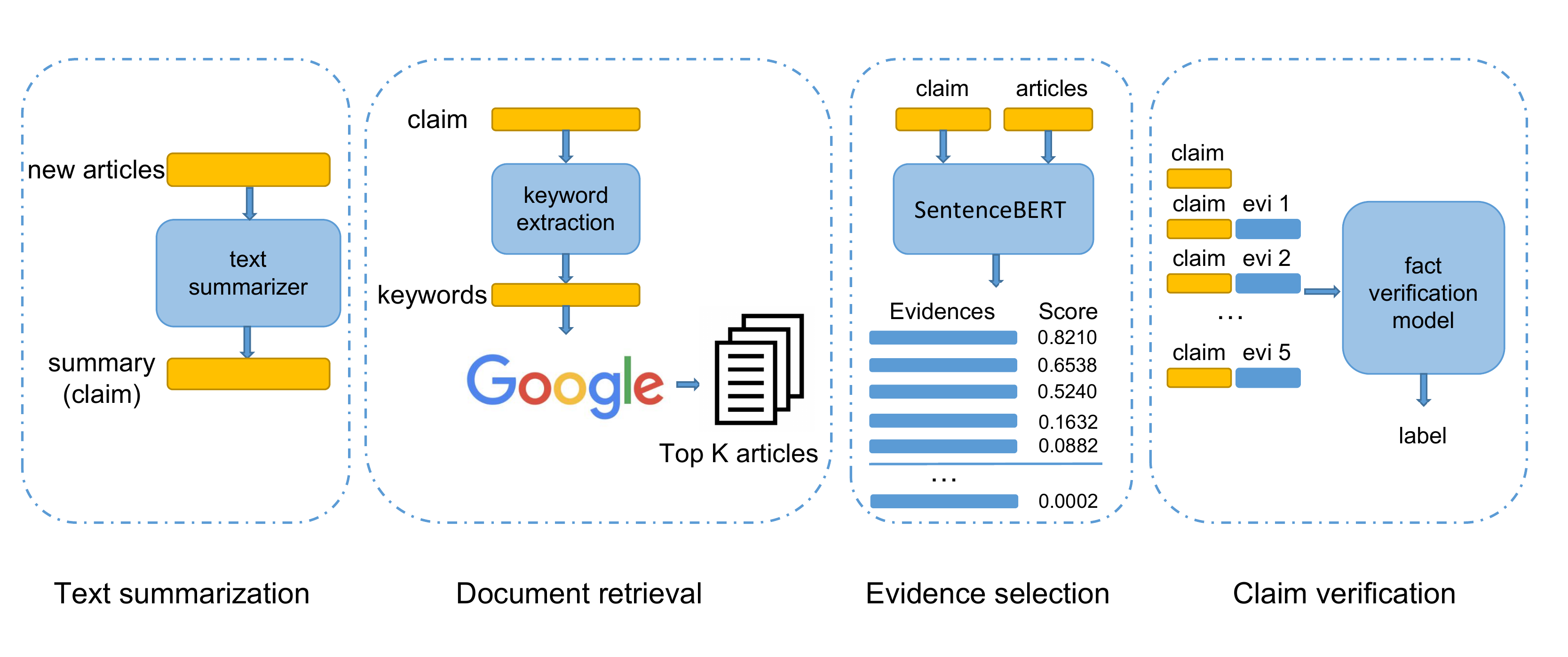}
    \caption{Illustration of the proposed approach. Our approach exploits two off-the-shelf models trained with text summarization and fact verification datasets, which are two relatively well-studied problems to perform zero-shot fake news detection.}
    \label{fig:illu}
\end{figure*}

To address the above issues, in this paper, we propose a simple yet effective approach to connect the dots between fact verification and fake news detection. Our approach exploits off-the-shelf models in two relatively well-studied problems---text classification and fact verification---to tackle the problem of fake news detection. Specifically, our approach first employs a text summarization model pre-trained on news corpora to summarize the long news article into a short claim. Then we use a fact verification model pre-trained on the FEVER dataset to detect whether the input news article is real or fake. Our approach transfers the recent success of fact verification models to enable zero-shot fake news detection, alleviating the requirement of large scale datasets to train fake news detection models. Experimental results on FakenewsNet, a benchmark dataset for fake news detection, show that our approach can achieve comparable performance with several competitive supervised content-based fake news detection models in a zero-shot fashion. In addition, when fine-tuning the fact verification model with labeled examples, our approach significantly outperforms the supervised baselines. This demonstrates the effectiveness of our proposed approach.

\section{Related Work}

The problem of fake news detection has gained much research interests with the widespread of fake news on social media~\cite{zhou2018fake}. Existing fake news detection models can be categorized into content-based models~\cite{conroy2015automatic,pan2018content} and social context based models~\cite{shu2017fake,qian2018neural,tschiatschek2018fake}. In this paper, we focus on content-based models that enable early fake news detection before the spread of fake news. Recent content-based fake news detection models~\cite{reddy2020text,shu2019defend,zhang2020fakedetector} generally formulates the problem as a text classification problem. However, many news articles are very long, which hinders the application of state-of-the-art pre-trained language models such as BERT~\cite{devlin2018bert} because their maximum context length is generally 512. Moreover, few large scale fake news detection datasets are available and they are generally limited in their domains while human annotation of fake news is time-consuming and expensive, which hinders the application of current supervised content-based fake news detection models. Another recent work related to our paper is an unsupervised fake news detection method proposed by~\cite{yang2019unsupervised}. However, their approach is based on the social context, which requires additional information and thus less general to content-based approaches. Our approach, in contrast, is the first fake news detection method that enables zero-shot fake news detection without any labeled examples or additional information to the best of our knowledge.

\section{Methodology}

Our approach is a novel content-based fake news detection method that exploits off-the-shelf pre-trained models in well-established text summarization and fact verification problems to enable zero-shot fake news detection without training on labeled datasets. As illustrated in Figure \ref{fig:illu}, in the first stage, our approach employs a text summarization model pre-trained on large scale news corpus to summarize the input long news article into a short claim which consists of one or a few sentences. Afterward, we use a fact verification model pre-trained on FEVER~\cite{thorne2018fever}, a large scale fact verification dataset, to check the trustworthiness of the article. We then describe the details of the two stages in our proposed approach.

For the first stage. we employ an open-sourced BERT-based extractive summarization model\footnote{https://github.com/dmmiller612/bert-extractive-summarizer}~\cite{miller2019leveraging} to summarize the input news article into a short claim. The length of the output summary is controlled to mimic that of the claims in the FEVER dataset by setting the compression ratio set to 0.1 and selecting the top 2 predicted sentence, which minimizes the inconsistency between the input of the fact verification model used in the second stage during its training and inference procedure. In addition, opinions in long articles are generally scattered, which makes it difficult for a fake news detection model to judge the reliability of long articles. Our approach makes the information more concentrated in a few sentences, making it easier to identify the trustworthiness of the article.

During the second stage, we employ GEAR~\cite{zhou2019gear}, a competitive fact verification model based on BERT and graph neural network trained on the FEVER dataset to verify the claim generated during the first stage. However, the model requires multiple support evidence, which is available in the FEVER dataset, to make the prediction. To mitigate this issue, we propose to construct evidence of the claim using a commercial search engine such as Google. Concretely, we first extract the keywords in the claim with AllenNLP\footnote{https://github.com/allenai/allennlp}~\cite{gardner2018allennlp}, an open-sourced NLP toolkit. Afterward, we feed the keywords into a search engine to crawl web texts related to the claim. We filter the urls that contains only pdf or images or the text content is less than 100 tokens, and select support evidence for the claim according to the sentence embedding similarity under SentenceBERT~\cite{reimers2019sentence}, a pre-trained sentence embedding model.  We select the top 5 related sentences as evidences and feed them into the pre-trained GEAR model together with the original claim to predict whether the claim is trustworthy or not. Note that the fact verification model pre-trained on FEVER is a three-way classification model including the ``Not Enough Infomation'' class. We simply omit this class and compare the output probability of the real and fake class to make the binary classification. 

In addition, when labeled examples are available, we can further fine-tune the fact verification model pre-trained on the FEVER dataset in a continual learning fashion with our target dataset to achieve better performance. For the experiments with fine-tuning the fact verification model with labeled fake news detection examples, we initialize the output layer to make the model perform binary classification and use the default hyperparameter of GEAR to perform the fine-tuning procedure.

\section{Experiments}

\subsection{Experimental Settings}

\paragraph{Dataset}

\begin{wraptable}{l}{5.5cm}
	\caption{Dataset Statistics.}
	\label{data}
		\begin{tabular}{lcc}
			\toprule
			\bf Domain & \bf \#Fake & \bf \#Real  \\
			\midrule
			\bf PolitiFact & 330 & 332 \\
			\bf GossipCop  & 4582 & 14477  \\
			\bottomrule
	\end{tabular}

\end{wraptable} 


We adopt FakenewsNet~\cite{shu2018fakenewsnet}, a recently released benchmark dataset for fake news detection in our experiments. The dataset consists of news articles in two domains including PolitiFact and GossipCop. We run the official code of the paper to obtain the datasets, detailed statistics of the dataset are presented in Table \ref{data}. We use 80\% of data for training and 20\% for testing. For evaluation metrics, we use accuracy, precision, recall, and F1 score following previous work~\cite{shu2018fakenewsnet}. In addition, we manually increase the ratio between real and fake news in the test set to simulate the real-world scenario where only a small portion of news articles are fake. Therefore, a random guessing baseline would achieve very poor results (i.e., around 0.2 to 0.3 F1 score). 

\paragraph{Compared Models} We adopt two settings of baseline to evaluate the effectiveness of the proposed approach. The first is the transfer learning setting where we assume that we have access to labeled examples in another domain but do not have training examples in the target domain. In this setting, we train a supervised fake news detection model with the labeled data in another domain and directly transfer the model for inference in our target domain without parameter updating. The second setting is the supervised setting where we train a fake news detection model in a supervised fashion with training examples in the target domain. We employ doc2vec~\cite{le2014distributed}, a pre-trained paragraph embedding approach to encode the news articles into a fixed-length vector with 300 dimension. We then apply standard machine learning models that yield state-of-the-art performance on the task of fake new detection including support vector machines
(SVM), logistic regression (LR), Naive Bayes (NB) as the classification model. Note that we do not compare against models based on pre-trained language models because the length of news articles are generally longer than their maximum context length and clipping the articles severely affects the model's performance. Conventional CNN or LSTM-based models compare similarly to machine learning based models in our preliminary experiments. We suspect this is because the pre-trained doc2vec is already a good feature extractor.

\subsection{Experimental Results}

\begin{table*}
	\centering
		\begin{tabular}{lcccccccc}
			\toprule
			\bf Model & \multicolumn{4}{c}{\bf PolitiFact} & \multicolumn{4}{c}{\bf GossipCop} \\
			& \bf Accuracy & \bf Precision & \bf Recall & \bf F1 & \bf Accuracy & \bf Precision & \bf Recall & \bf F1 \\
			\midrule 
			\multicolumn{9}{c}{\textbf{Zero-Shot Setting}} \\
			\midrule
			\textbf{SVM} &33.94  & 22.31& 16.97& 19.27&44.45 & 23.71 & \bf 55.10 & 33.15 \\
			\textbf{NB} & 37.74& 26.47& 19.09 & 22.18  & 47.19 & 24.40 & 53.02 & 33.42 \\
			\textbf{LR} &35.49& 27.30& 23.33 & 25.16  & 49.09 & 23.80 & 47.08 &31.61 \\
			\textbf{Ours} & \bf 44.10 & \bf 44.82 & \bf 52.42 & \bf 48.32& \bf 56.49 & \bf 29.09 & 52.42 & \bf 37.42 \\
			\midrule
			\multicolumn{9}{c}{\textbf{Supervised Setting}} \\
			\midrule
			\textbf{SVM} & 49.30 & 49.27 & 47.22 & 48.23 & 52.71 & 28.53 & 58.27 & 38.30 \\
			\textbf{NB} & 59.72 & 60.93 & 54.17 & 57.35 & 62.10 & 32.58 & 48.23 & 38.89\\
			\textbf{LR} & 49.30 & 49.20 & 43.05 & 45.92 & 57.47 & 31.08 & 57.60 & 40.37\\
			\textbf{Ours} & \bf 68.75 & \bf 65.17  & \bf 80.56 & \bf 72.50 & \bf 73.74 & \bf 47.64 & \bf 58.77 & \bf 52.50 \\
			\bottomrule
	\end{tabular}
	\caption{Test set performance of compared models on two domains in the FakenewsNet benchmark in both the zero-shot and the supervised setting. For the baseline methods in the zero-shot setting, we train a supervised model on the training set of one domain}
	\label{mainresult}
\end{table*}

We present the experimental results in both zero-shot and supervised setting in Table \ref{mainresult}. We can see that the zero-shot variant of our approach significantly outperforms all compared zero-shot baselines, demonstrating that our approach can successfully connect the dots between a well-trained fact verification model and the task of fake news detection. It is notable that our approach does not require any fake news detection training example while the zero-shot baselines require first training on a fake news detection model in a different domain. It is also remarkable that our approach yields comparable results  compared to the baselines trained in the supervised setting. When fine-tuning the fact verification model used in our approach with labeled training examples, our approach outperforms the supervised baselines with a large margin (i.e., over 10 F1 scores). This shows that our approach is also helpful when training examples are available.


\subsection{Analysis}

\begin{wrapfigure}{l}{3.5cm}
\vspace{-0.3cm}
\includegraphics[width=3.5cm]{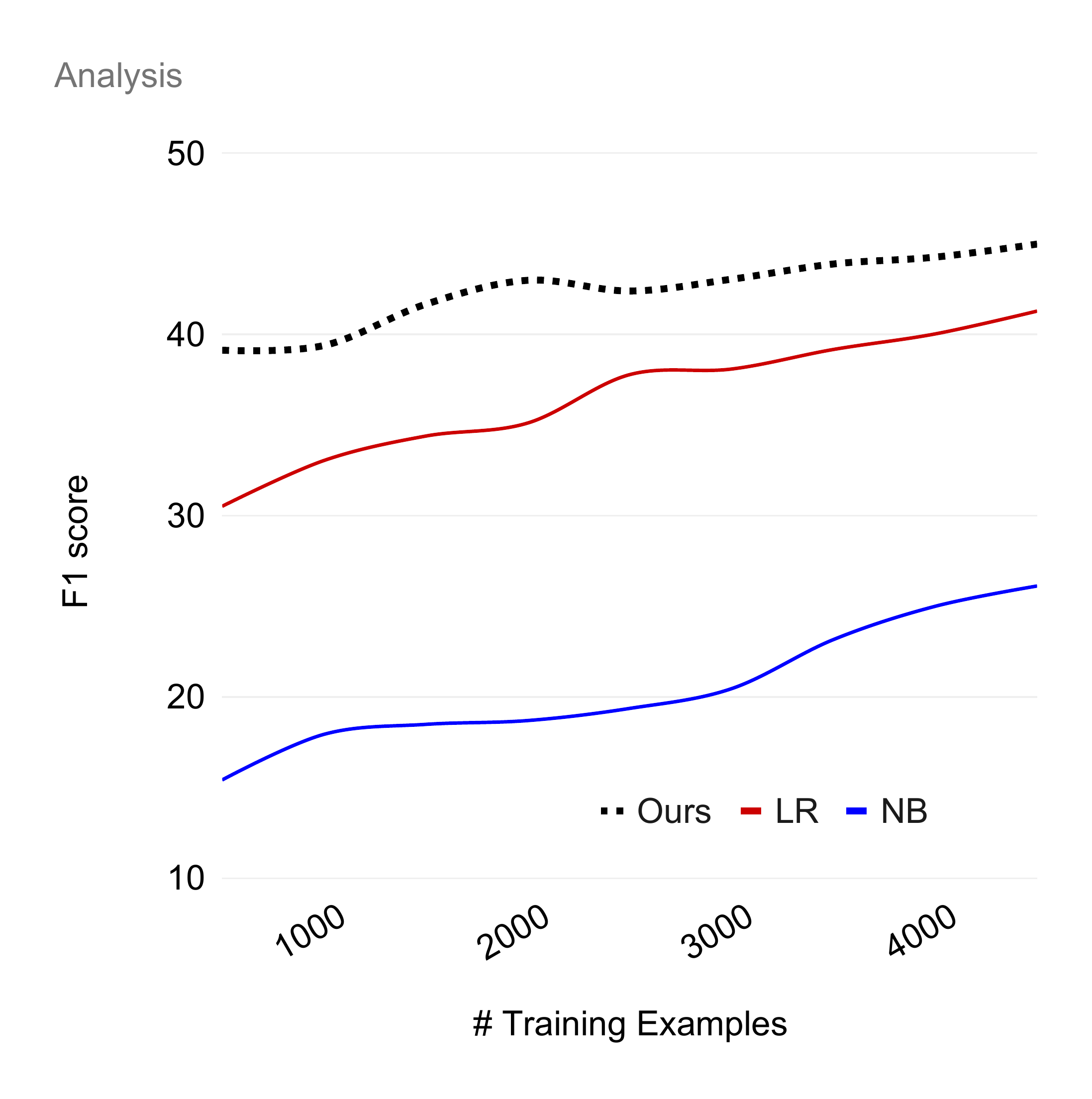}
\caption{Analysis.}\label{fig:analysis}
\end{wrapfigure}

To better understand our proposed approach, we conduct an analysis investigating the effectiveness of transferring a well-trained fact verification model for fake news detection. Specifically, we fine-tune the fact verification model with different numbers of training examples to simulate the transition from the zero-shot setting to the supervised setting and compare it to the supervised baselines with the same range of training examples.

The results are presented in Figure \ref{fig:analysis}. We can see that the performance yielded by our approach is already relatively good with only 500 to 2000 training examples. In contrast, the compared supervised baselines need more training examples to achieve comparable performance. This confirms the effectiveness of our approach to reduce the need of large scale training data for training fake news detection models.

\section{Discussion \& Conclusion}
In this paper, we propose a novel fake news detection approach that exploits well-trained text summarization model and fact verification model. By connecting the dots between fact verification and fake news detection, our approach enables zero-shot fake news detection. There exists previous work that exploit models trained on one task for another related task, such as using machine translation models for paraphrasing~\cite{somers2005round} and grammatical error correction~\cite{lichtarge2019corpora,zhou2019improving}. Our approach is similar to this line of research but combines two models in other tasks for another downstream task. 

Experiments on a fake news detection benchmark show that our approach can yield comparable performance with competitive supervised content-based fake news detection models in a zero-shot fashion. Our approach can also leverage labeled examples more effectively than conventional supervised methods with continual learning, enabling building fake news detection models more effectively for domains where few labeled data are available. One limitation of our approach is that it is based on pre-trained language models and thus can be computationally expensive to use in real-world applications. Therefore, for future work, we plan to exploit methods to accelerate the inference of pre-trained language models including model compression methods like DistilBERT~\cite{sanh2019distilbert} and BERT-of-Theseus~\cite{xu2020bert}, as well as adaptive inference methods such as PABEE~\cite{zhou2020bert}.

\section*{Acknowledgments}
We thank the anonymous reviewers for their valuable comments.

\bibliographystyle{coling}
\bibliography{coling2020}

\end{document}